\documentclass{article}

\usepackage[final]{neurips_2021}

\usepackage{amsmath}
\usepackage{amssymb}
\usepackage{diagbox}
\usepackage{epsfig}
\usepackage{graphicx}
\usepackage{wrapfig}
\usepackage{caption}
\usepackage{natbib}
\setcitestyle{square,numbers}

\usepackage[utf8]{inputenc} 
\usepackage[T1]{fontenc}    
\usepackage{hyperref}       
\usepackage{url}            
\usepackage{booktabs}       
\usepackage{amsfonts}       
\usepackage{nicefrac}       
\usepackage{microtype}      
\usepackage{xcolor}         

\title{Highly Efficient Representation and Active Learning Framework and Its Application to Imbalanced Medical Image Classification}

\author{
  Heng Hao \thanks{\texttt{h.heng@samsung.com}} \qquad
  Hankyu Moon \qquad
  Sima Didari \qquad
  Jae Oh Woo \qquad
  Patrick Bangert\\ 
  Samsung SDS Research America\\}

\begin{document}

\maketitle

\begin{abstract}
		We propose a highly data-efficient active learning framework for image classification. Our novel framework combines: (1) unsupervised representation learning of a Convolutional Neural Network and (2) the Gaussian Process (GP) method, in sequence to achieve highly data and label efficient classifications. Moreover, both elements are less sensitive to the prevalent and challenging class imbalance issue, thanks to the (1) feature learned without labels and (2) the Bayesian nature of GP. The GP-provided uncertainty estimates enable active learning by ranking samples based on the uncertainty and selectively labeling samples showing higher uncertainty. We apply this novel combination to the severely imbalanced case of COVID-19 chest X-ray classification and the Nerthus colonoscopy classification. We demonstrate that only $\lesssim 10\%$ of the labeled data is needed to reach the accuracy from training all available labels. We also applied our model architecture and proposed framework to a broader class of datasets with expected success.
\end{abstract}

\section{Introduction}
Medical imaging is one of the major applications of computer vision technologies. The applications range from the most straightforward task of image classification (such as X-Ray, ultrasound, fundus) to image segmentation (anatomy or legions), 3D imaging, and functional imaging (fMRI). Our focus in this paper is image classification and its applications.

There has been significant progress in the past decade both in terms of theoretical insights and classification accuracy due to the development and adoption of Deep Neural Network (DNN) models. The most well-established approach is the supervised training of Convolutional Neural Networks (CNN), which first identifies informative image features from multiple layers of convolutional filters fed to a small number of classification layers that produce category decisions. However, this popular approach typically requires large numbers of labeled images from each category to achieve an accuracy level useful for medical diagnosis. Data collection and labeling are often very costly. In some cases, it is not feasible to collect enough data for a quick automated diagnosis, as experienced in the time-critical cases of the COVID-19 pandemic. This leads to a highly imbalanced class distribution (``Normal" cases \(\gg\) ``COVID-19" cases) that negatively impacts the decision accuracy. Given these practical challenges, we depart from the standard approach and propose a highly data-efficient methodology that can achieve the same level of accuracy using significantly fewer images and labels. It is based on CNN unsupervised representation learning hybrid with a Gaussian Process (GP) classifier. The GP-provided uncertainty estimates enable active learning by ranking unlabelled samples and selectively labeling samples showing higher uncertainty.

\section{Methodology}

\begin{figure}[t]
	\centering
	\includegraphics[width=0.9\linewidth]{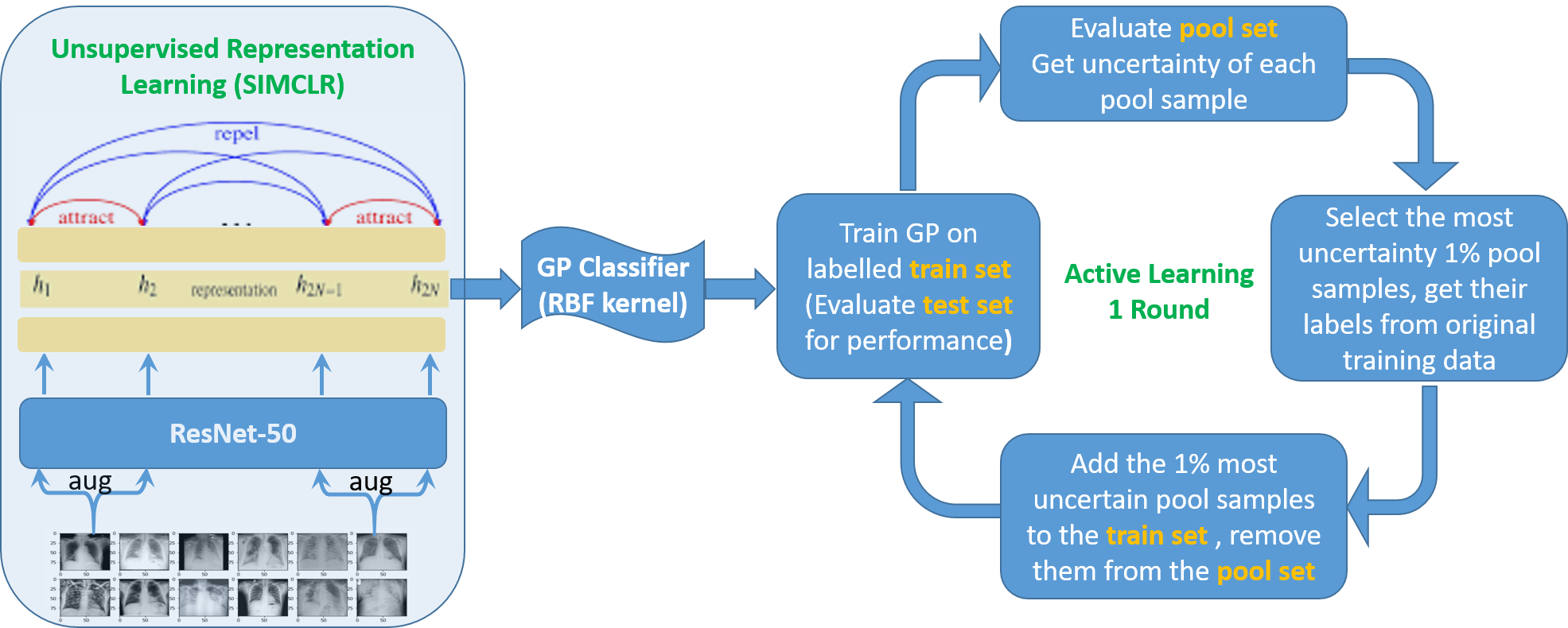}
	\caption{Our active learning framework is illustrated. The representation generator is trained unsupervised with contrastive loss. The representations are used as inputs to the GP classifier. The GP classifier is trained in the active learning loop until the target performance is reached. 
		\label{fig:framework}}
	\vspace{-1em}
\end{figure}
\textbf{Active learning} \cite{cohn1996active} is one of the most powerful techniques to improve data efficiency by saving labeling efforts. Its primary goal is to use the minimum amount of labels to reach maximum performance. We start with a small amount of labeled data (initial train set) to train the model. Then, we use an \textit{acquisition function} (often based on the prediction uncertainty or entropy) to evaluate the unlabeled pool set, choose the most helpful pooling samples, and ask the external \textit{oracle} (generally a human) for the label. These newly labeled data are then added to the train set to update the model. This process is repeated multiple times, with the train set gradually increasing in size until the model performance (evaluate with the holding test set) reaches a particular stopping criterion. Active learning will considerably facilitate real-world adoption of AI \cite{adomavicius2005toward, calinon2007onlearning, hoi2006albatch, siddhant2018deep, tong2001active}, especially in medical imaging where data collection and labeling are quite expensive.

The \textbf{CNN-GP hybrid model} is trained in two decoupled steps, following the practical guidance regarding the benefits of decoupling \cite{kang2019decoupling, tang2020long}. We illustrate our training framework in Figure~\ref{fig:framework}. 

The first step is the \textbf{representation learning}, which refers to an unsupervised training step with the goal of extracting image features that are used in diverse downstream tasks. Among many different approaches for this challenging task, the {\it{contrastive loss}} based learning has been applied very successfully and shows state-of-the-art performance in classification~\cite{oord2018representation, bachman2019learning, zhuang2019local, caron2020unsupervised, he2020momentum, chen2020simple, chen2020big, grill2020bootstrap, li2020prototypical}. Especially Chen et al.~\cite{chen2020big} confirms very high label-efficiency of the learned representation. These results have clear implications to data imbalance problem: the learned features (1) do not overfit dominant classes because the training does not use the class information (2) capture less dominant classes more efficiently. 

We start with a mini-batch with $N=16$ image samples, image augmentation (random crop, random flip, color distortion, Gaussian blur, and random gray-scale) is applied to each image twice to generate an image pair, leading to total $2N$ samples.  The training maximizes the similarity of the positive pair (the ones augmented from the same image), leveraging a contrastive loss. In SimCLR method \cite{chen2020simple, chen2020big}, the contrastive loss is evaluated at the projection head layer after the ResNet-50 backbone. 

The second step is the \textbf{Gaussian Process (GP)} classifier, a non-paramtric Bayesian method, that can produce the prediction and its uncertainty in one shot. Many techniques have been developed to extract Bayesian uncertainty estimates from DNN~\cite{gal16dropout, graves2011practical}, and it was observed that the lower layers of a DNN for images may not benefit as much from a Bayesian treatment~\cite{kendall2017bayesian}. Similarly, GP has been used at the top of DNNs and has been applied to both classification and regression problems while producing uncertainty estimates~\cite{calandra2016manifold,bradshaw2017adversarialeu}. However, it was observed that a DNN such as ResNet efficiently learns CIFAR10 with a test accuracy of more than 96\%, while kernel methods such as GP can barely reach 80\% \cite{wilson2016stochastic}. It is well accepted \cite{allenzhu2019whatcr} that better representation input to the kernel is crucial. The contrastive learning representation described above, which aims to aggregate points with similarity measured by distance, will be the ideal input to a GP with a distance-based RBF (radial basis function) kernel \footnote{The RBF kernel is  $k(r)=\sigma^2\exp(-r^2/2l^2)$, where $r$ is the Euclidean distance between input vectors, $l$ length-scaler and $\sigma^2$ variance-scaler are two hyper-parameters.}. This decoupled two-step training CNN-GP hybrid model framework, not only alleviates the inline training issue of GP but also extracts a predetermined lower dimensional feature space for the GP, so that the GP classifier shows accuracy advantage over the Bayesian linear classifier, or the finite width non-linear neural network \cite{neal1996bayesian,bui2016deep}. The GP classifier also offers two special properties that make it well-suited for medical image analysis: (1) As a Bayesian method, it provides the prediction and its uncertainty in one-shot, which will greatly help medical diagnosis; (2) Compared to other non-Bayesian methods, it handles the issue of class imbalance more effectively \cite{rosevear2017gaussian}, which is quite common in medical data.

Once we calculate the representations (2,048 dimensions for each sample) of all the train set, we use them as the input to the GP classifier. To train the multi-class GP, we use the Sparse Variational GP (SVGP) \cite{hensman2013gaussian} from GPflow package \cite{GPflow2017}. We choose the RBF kernel with 128 inducing points. We trained the model for 24 epochs using Adam optimizer with a learning rate of 0.001. The one-shot output of the GP classifier include both the mean and variance of the class probability. We calculate the mean variance of the class probability as prediction uncertainty for all the pool data and select the next batch of images showing the largest uncertainties.

\section{Experimental Results}\label{sec:experiments}
\subsection{COVID-19 Dataset and Active Learning Results}\label{sec:covid}
\begin{wrapfigure}{l}{0.6\textwidth}
	\centering
	\vspace{-1em}
	\includegraphics[width=.50\textwidth]{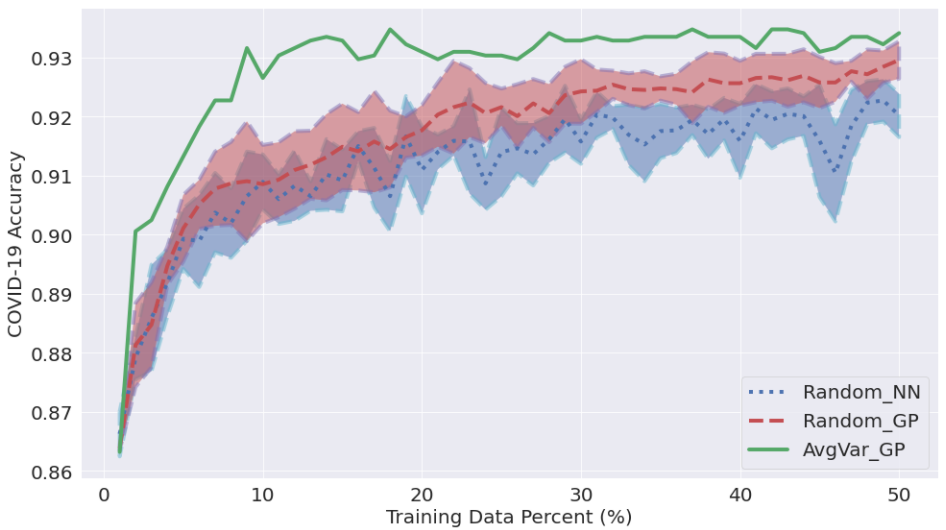}
	\caption{Test accuracy from active learning compared to random selection for COVIDx dataset. The line and shaded area show the mean and standard deviation of five runs. \label{fig:al0}}
	\vspace{-1em}	
\end{wrapfigure}

COVIDx dataset is the largest open access benchmark dataset in terms of the number of COVID-19 positive patient cases to the best of our knowledge. At the time of this study, it consisted of 15,521 chest X-ray images, of which 8,851 are ``Normal", 6,063 ``Pneumonia" and 601 ``COVID-19" pneumonia cases. The dataset is the combination and modification of five different publicly available data repositories. These datasets are: (1) COVID-19 Image Data Collection \cite{cohen2020covid19}, (2) COVID-19 Chest X-ray Dataset Initiative  \cite{Chung2020} (3) ActualMed COVID-19 Chest X-ray Dataset Initiative \cite{Chung2020_2} (4) RSNA Pneumonia Detection Challenge Dataset, which is a collection of publicly available X-rays  \cite{RSNA2019} , and (5) COVID-19 Radiography Database  \cite{COVID19}. The dataset is highly imbalanced with significantly fewer COVID positive cases than other conditions: the train sample (7966 ``Normal" , 5469 ``Pneumonia", and 507 ``COVID-19") and the test sample (885 ``Normal", 594 ``Pneumonia", and 100 ``COVID-19"). About \text{4\%} are COVID-19 positive cases in the train sample.

Before feeding data to the representation generator, we pre-process the images by performing a 15\% top crop, re-centering, and resizing to the original image size to delete the embedded textual information and enhance the region of interest \cite{maguolo2020critic, tartaglione2020unveiling}. 

In Figure~\ref{fig:al0}, we compare the CNN-GP hybrid active learning with the uncertainty acquisition function (green) versus the same model but random selection (red), and the same SIMCLR backbone with softmax layers model with random selection (blue) to show the benefit of both the active learning and GP. We start with the same initial batch for a fair comparison. To check the consistency of our results, we repeat multiple (five) random runs. The lines and shaded area shown in the figure are the means and standard deviation of the five independent runs for random selection respectively. With the unsupervised representation learning followed by a GP classifier, only $\sim 10\%$ of the training data needs to be labeled to achieve the same accuracy as if all the labeled training data is used. Especially when the sample size is small ($<20\%$), the training data selected by the acquisition functions accelerates the model to reach significantly higher test accuracy. The remaining 90\% of the data offer no new information to the classification model and can be auto-labeled by the CNN-GP hybrid model, saving considerable labeling cost.

\begin{wraptable}{l}{0.46\textwidth}
	\begin{tabular}{||c || c | c ||} 
		\hline 
		Fractions & Pneumonia & COVID-19 \\
		\hline\hline
		\textbf{Train} & \textbf{39.22\%} & \textbf{3.64\%} \\ 
		\hline\hline
		batch 1 & 47.86\% & 25.71\% \\
		\hline
		batch 2 & 47.14\% & 10.71\% \\
		\hline
		batch 3 & 44.29\% & 9.71\% \\
		\hline
		batch 4 & 47.86\% & 3.57\% \\
		\hline
		batch 5 & 44.29\% & 7.14\% \\
		\hline
		batch 6 & 47.14\% & 10.00\% \\
		\hline
		batch 7 & 42.14\% & 11.43\% \\
		\hline
		batch 8 & 47.14\% & 7.86\% \\
		\hline
	\end{tabular}
	\captionof{table}{Fraction of ``Pneumonia" and ``COVID-19" in each batch of the active learning cycles compared to the whole train set (random selection).}
	\label{tab:alsel}
	\vspace{-2em}
\end{wraptable}

Even with the same random selection, the CNN-GP hybrid model are generally more stable (with tighter shaded area) and performs better (mean accuracy curve is higher) than the SIMCLR CNN with a softmax layer.

The labels of the samples selected in each cycle based on the uncertainty acquisition function is checked in Table~\ref{tab:alsel}. The samples selected by active learning has much more fraction of ``Pneumonia" and ``COVID-19" samples compared to the whole train set (random selection). It is clearly shown that more samples that belong to the rare classes are automatically selected in the early active learning cycles, showing the proposed active learning framework is more robust towards the class imbalance issue. 

\subsection{Nerthus Dataset and Active Learning Results}\label{sec:Ncolon}

\begin{wraptable}{l}{0.46\textwidth}
	\begin{tabular}{||c || c c c c||} 
		\hline
		& 0 & 1 & 2 & 3 \\  
		\hline\hline
		Train & 447 & 2434 & 867 & 1252 \\ 
		\hline
		Test & 13 & 82 & 33 & 48 \\
		\hline
	\end{tabular}
	\captionof{table}{Nerthus Dataset}
	\label{tab:datasplit}
	\vspace{-1em}
\end{wraptable}

The Nerthus dataset \cite{pogorelov2017nerthus} we use contains 5176 frames of colonoscopy images from 21 videos. We randomly select 176 frames as test set and the rest 5000 as train and pool set. The class distribution are depicted in Table~\ref{tab:datasplit}. We can clearly see that the Nerthus data is more balanced compared to the Covid-19 data. We also perform necessary cleaning to remove the unrelated tagging regions in the images before feeding into the deep learning model.

\begin{wrapfigure}{l}{0.6\textwidth}
	\centering
	\vspace{-1em}
	\includegraphics[width=.50\textwidth]{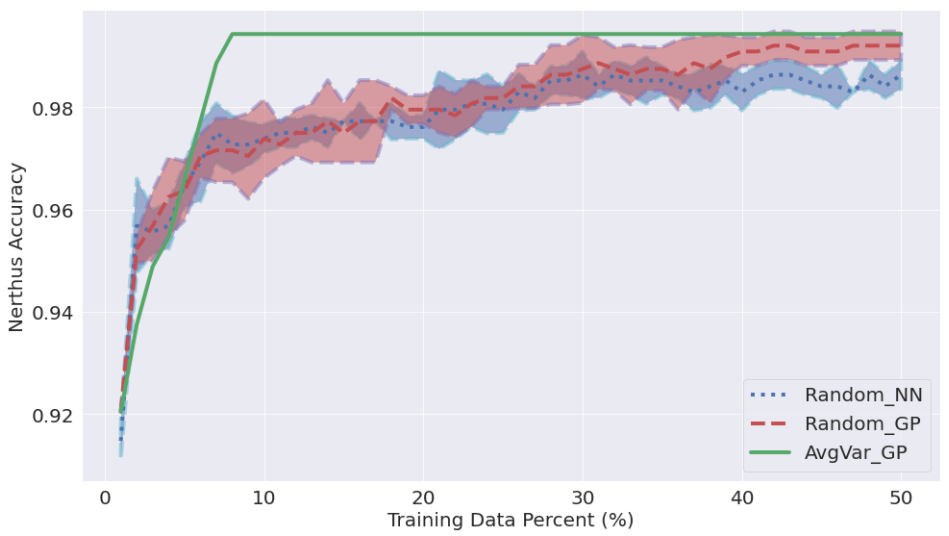}
	\caption{Test accuracy from active learning compared to random selection for Nerthus dataset. \label{fig:al1}}
	\vspace{-1em}	
\end{wrapfigure}

We perform same experiment for the Nerthus dataset as Section~\ref{sec:covid}. The result is shown in Figure~\ref{fig:al1}. With the CNN-GP hybrid model with uncertainty acquistion function, only $8\%$ of the training data needs to be labeled to achieve the same accuracy as if all the labeled training data is used. When compare the CNN-GP hybrid model with the CNN with softmax layer model, we got similar results that is the CNN-GP hybrid model is more stable and more accurate even with the same CNN backbone. 

\section{Conclusion}
We introduced a data-efficient CNN-GP hybrid model and showed that our approach enables an efficient CNN-GP active learning with its application to the highly imbalanced COVID-19 chest X-ray imaging and Nerthus colonoscopy images, leading to saving $\sim90\%$ of the labeling time and cost. Using the uncertainty generated from the GP model as acquisition function tends to select more less represented class samples in the early stages of the active learning cycles. 

We applied the proposed framework in several other datasets and reaches expected success as the above two examples. Further improvement of the proposed framework is attainable through improved unsupervised representation learning and implementation of better acquisition functions with stronger exploration and exploitation characteristics. The aforementioned directions will be the focus of our future studies.


{\small
\bibliographystyle{plain}
\bibliography{GPAL}}

\begin{thebibliography}{10}

\bibitem{adomavicius2005toward}
Gediminas Adomavicius and Alexander Tuzhilin.
\newblock Toward the next generation of recommender systems: A survey of the
  state-of-the-art and possible extensions.
\newblock {\em IEEE Transactions on Knowledge \& Data Engineering}, 2005.

\bibitem{allenzhu2019whatcr}
Zeyuan Allen-Zhu and Y.~Li.
\newblock What can resnet learn efficiently, going beyond kernels?
\newblock {\em ArXiv}, abs/1905.10337, 2019.

\bibitem{bachman2019learning}
Philip Bachman, R~Devon Hjelm, and William Buchwalter.
\newblock Learning representations by maximizing mutual information across
  views.
\newblock In {\em Advances in Neural Information Processing Systems}, pages
  15535--15545, 2019.

\bibitem{bradshaw2017adversarialeu}
John Bradshaw, A.~Matthews, and Zoubin Ghahramani.
\newblock Adversarial examples, uncertainty, and transfer testing robustness in
  gaussian process hybrid deep networks.
\newblock {\em arXiv preprint arXiv:1707.02476}, 2017.

\bibitem{bui2016deep}
Thang Bui, Daniel Hernandez-Lobato, Jose Hernandez-Lobato, Yingzhen Li, and
  Richard Turner.
\newblock Deep gaussian processes for regression using approximate expectation
  propagation.
\newblock In {\em Proceedings of The 33rd International Conference on Machine
  Learning}, volume~48 of {\em Proceedings of Machine Learning Research}, pages
  1472--1481, New York, New York, USA, 20--22 Jun 2016. PMLR.

\bibitem{calandra2016manifold}
R.~{Calandra}, J.~{Peters}, C.~E. {Rasmussen}, and M.~P. {Deisenroth}.
\newblock Manifold gaussian processes for regression.
\newblock In {\em 2016 International Joint Conference on Neural Networks
  (IJCNN)}, pages 3338--3345, 2016.

\bibitem{calinon2007onlearning}
Sylvain Calinon, Florent Guenter, and Aude Billard.
\newblock On learning, representing, and generalizing a task in a humanoid
  robot.
\newblock {\em IEEE Transactions on Systems, Man and Cybernetics, Part B
  (Cybernetics)}, 37(2):286--298, 2007.

\bibitem{caron2020unsupervised}
Mathilde Caron, Ishan Misra, Julien Mairal, Priya Goyal, Piotr Bojanowski, and
  Armand Joulin.
\newblock Unsupervised learning of visual features by contrasting cluster
  assignments.
\newblock {\em Advances in Neural Information Processing Systems}, 33, 2020.

\bibitem{chen2020simple}
Ting Chen, Simon Kornblith, Mohammad Norouzi, and Geoffrey Hinton.
\newblock A simple framework for contrastive learning of visual
  representations.
\newblock {\em Proceedings of the International Conference on Machine
  Learning(ICML)}, 2020.

\bibitem{chen2020big}
Ting Chen, Simon Kornblith, Kevin Swersky, Mohammad Norouzi, and Geoffrey~E
  Hinton.
\newblock Big self-supervised models are strong semi-supervised learners.
\newblock {\em Advances in Neural Information Processing Systems}, 33, 2020.

\bibitem{Chung2020_2}
A~Chung.
\newblock Actualmed covid-19 chest x-ray data initiative.
\newblock \url{https://github.com/agchung/Actualmed-COVID-chestxray-dataset},
  2020.

\bibitem{Chung2020}
A~Chung.
\newblock Figure1-covid-chestxray-dataset.
\newblock \url{ https://github.com/agchung/Figure1-COVID-chestxray-dataset},
  2020.

\bibitem{cohen2020covid19}
Joseph~Paul Cohen, Paul Morrison, and Lan Dao.
\newblock Covid-19 image data collection, 2020.

\bibitem{cohn1996active}
David~A. Cohn, Zoubin Ghahramani, and Michael~I. Jordan.
\newblock Active learning with statistical models.
\newblock {\em Journal of Artificial Intelligence Research}, 4:129--145, 1996.

\bibitem{gal16dropout}
Yarin Gal and Zoubin Ghahramani.
\newblock Dropout as a bayesian approximation: Representing model uncertainty
  in deep learning.
\newblock In {\em Proceedings of The 33rd International Conference on Machine
  Learning}, pages 1050--1059, 2016.

\bibitem{graves2011practical}
Alex Graves.
\newblock Practical variational inference for neural networks.
\newblock {\em Advances in Neural Information Processing System},
  24:2348--2356, 2011.

\bibitem{grill2020bootstrap}
Jean-Bastien Grill, Florian Strub, Florent Altch{\'e}, Corentin Tallec, Pierre
  Richemond, Elena Buchatskaya, Carl Doersch, Bernardo Avila~Pires, Zhaohan
  Guo, Mohammad Gheshlaghi~Azar, et~al.
\newblock Bootstrap your own latent-a new approach to self-supervised learning.
\newblock {\em Advances in Neural Information Processing Systems}, 33, 2020.

\bibitem{he2020momentum}
Kaiming He, Haoqi Fan, Yuxin Wu, Saining Xie, and Ross Girshick.
\newblock Momentum contrast for unsupervised visual representation learning.
\newblock In {\em Proceedings of the IEEE/CVF Conference on Computer Vision and
  Pattern Recognition}, pages 9729--9738, 2020.

\bibitem{hensman2013gaussian}
James Hensman, Nicoló Fusi, and Neil~D. Lawrence.
\newblock Gaussian processes for big data.
\newblock In {\em Proceedings of the Twenty-Ninth Conference on Uncertainty in
  Artificial Intelligence (UAI2013)}, 2013.

\bibitem{hoi2006albatch}
Steven~CH Hoi, Rong Jin, Jianke Zhu, and Michael~R Lyu.
\newblock Batch mode active learning and its application to medical image
  classification.
\newblock In {\em Proceedings of the 23rd international conference on Machine
  Learning}, pages 1492--1501, 2016.

\bibitem{COVID19}
Kaggle.
\newblock Radiological society of north america. covid-19 radiography database.
\newblock
  \url{https://www.kaggle.com/tawsifurrahman/covid19-radiography-database},
  2019.

\bibitem{kang2019decoupling}
Bingyi Kang, Saining Xie, Marcus Rohrbach, Zhicheng Yan, Albert Gordo, Jiashi
  Feng, and Yannis Kalantidis.
\newblock Decoupling representation and classifier for long-tailed recognition.
\newblock In {\em International Conference on Learning Representations}, 2019.

\bibitem{kendall2017bayesian}
Alex Kendall, Vijay Badrinarayanan, and Roberto Cipolla.
\newblock Bayesian segnet: Model uncertainty in deep convolutional
  encoder-decoder architectures for scene understanding.
\newblock In Gabriel~Brostow Tae-Kyun~Kim, Stefanos~Zafeiriou and Krystian
  Mikolajczyk, editors, {\em Proceedings of the British Machine Vision
  Conference (BMVC)}, pages 57.1--57.12. BMVA Press, September 2017.

\bibitem{li2020prototypical}
Junnan Li, Pan Zhou, Caiming Xiong, Richard Socher, and Steven~CH Hoi.
\newblock Prototypical contrastive learning of unsupervised representations.
\newblock {\em arXiv preprint arXiv:2005.04966}, 2020.

\bibitem{maguolo2020critic}
Gianluca Maguolo and Loris Nanni.
\newblock A critic evaluation of methods for covid-19 automatic detection from
  x-ray images.
\newblock {\em arXiv preprint arXiv:2004.12823}, 2020.

\bibitem{GPflow2017}
Alexander G. de~G. Matthews, Mark {van der Wilk}, Tom Nickson, Keisuke. Fujii,
  Alexis {Boukouvalas}, Pablo {Le{\'o}n-Villagr{\'a}}, Zoubin Ghahramani, and
  James Hensman.
\newblock {{GP}flow: A {G}aussian process library using {T}ensor{F}low}.
\newblock {\em Journal of Machine Learning Research}, 18(40):1--6, apr 2017.

\bibitem{neal1996bayesian}
Radford~M. Neal.
\newblock {\em Bayesian Learning for Neural Networks}.
\newblock Springer-Verlag, Berlin, Heidelberg, 1996.

\bibitem{oord2018representation}
Aaron van~den Oord, Yazhe Li, and Oriol Vinyals.
\newblock Representation learning with contrastive predictive coding.
\newblock {\em arXiv preprint arXiv:1807.03748}, 2018.

\bibitem{pogorelov2017nerthus}
Konstantin Pogorelov, Kristin~Ranheim Randel, Thomas de~Lange, Sigrun~Losada
  Eskeland, Carsten Griwodz, Dag Johansen, Concetto Spampinato, Mario Taschwer,
  Mathias Lux, Peter~Thelin Schmidt, Michael Riegler, and P{\aa}l Halvorsen.
\newblock Nerthus: A bowel preparation quality video dataset.
\newblock In {\em Proceedings of the 8th ACM on Multimedia Systems Conference},
  MMSys'17, pages 170--174, New York, NY, USA, 2017. ACM.

\bibitem{rosevear2017gaussian}
David Rosevear and Alta de~Waal.
\newblock Gaussian processes applied to class-imbalanced datasets.
\newblock 12 2017.

\bibitem{RSNA2019}
RSNA.
\newblock Radiological society of north america. rsna pneumonia detection
  challenge.
\newblock
  \url{https://www.kaggle.com/c/rsna-pneumonia-detection-challenge/data}, 2019.

\bibitem{siddhant2018deep}
Aditya Siddhant and Zachary~Chase Lipton.
\newblock Deep bayesian active learning for natural language processing:
  Results of a large-scale empirical study.
\newblock {\em ArXiv}, abs/1808.05697, 2018.

\bibitem{tang2020long}
Kaihua Tang, Jianqiang Huang, and Hanwang Zhang.
\newblock Long-tailed classification by keeping the good and removing the bad
  momentum causal effect.
\newblock {\em Advances in Neural Information Processing Systems}, 33, 2020.

\bibitem{tartaglione2020unveiling}
Enzo Tartaglione, Carlo~Alberto Barbano, Claudio Berzovini, Marco Calandri, and
  Marco Grangetto.
\newblock Unveiling covid-19 from chest x-ray with deep learning: a hurdles
  race with small data.
\newblock {\em arXiv preprint arXiv:2004.05405}, 2020.

\bibitem{tong2001active}
Simon Tong.
\newblock {\em Active learning: theory and applications}.
\newblock PhD thesis, Stanford University, 2001.

\bibitem{wilson2016stochastic}
Andrew~G Wilson, Zhiting Hu, Ruslan~R Salakhutdinov, and Eric~P Xing.
\newblock Stochastic variational deep kernel learning.
\newblock In {\em Advances in Neural Information Processing Systems}, pages
  2586–--2594, 2016.

\bibitem{zhuang2019local}
Chengxu Zhuang, Alex~Lin Zhai, and Daniel Yamins.
\newblock Local aggregation for unsupervised learning of visual embeddings.
\newblock In {\em Proceedings of the IEEE International Conference on Computer
  Vision}, pages 6002--6012, 2019.

\end{thebibliography}

\end{document}